\documentclass{article}

 \usepackage[preprint]{neurips_2026}

\usepackage[utf8]{inputenc} 
\usepackage[T1]{fontenc}    
\usepackage{hyperref}       
\usepackage{url}            
\usepackage{booktabs}       
\usepackage{amsfonts}       
\usepackage{nicefrac}       
\usepackage{xcolor}         
\usepackage{enumitem}
\usepackage{graphicx}
\usepackage{booktabs}
\usepackage{multirow}
\usepackage{bbm}

\usepackage{amsmath}
\usepackage{caption}
\usepackage{wrapfig}
\usepackage{subcaption}
\usepackage{cleveref}
\usepackage{amsmath}                 
\usepackage{amssymb}                 
\usepackage{xspace}  
\usepackage[table]{xcolor}         

\usepackage{multirow}

\newcommand{\ourmethod}{\textit{DiDE}\xspace}
\newcommand{\ourproblem}{\textit{Disen3D}\xspace}
\newcommand{\ourbench}{{Disen3D-Bench}\xspace}
\newcommand{\minisection}[1]{\vspace{0.005in} \noindent {\bf #1}}

\title{\ourmethod:\textit{D}irect \textit{I}njection with Color-Texture \textit{DE}coupling for 3D Stylization
} 

\author{
\bfseries Tao Wu$^{1,2}$ \quad
Alexandra Gomez-Villa$^{1,2}$
\\
\bfseries Senmao Li$^{3}$ \quad
Yaxing Wang$^{4}$ \quad
Joost van de Weijer$^{1,2}$ \quad
Kai Wang$^{5,6}$\thanks{Corresponding author.}
\\[6pt]
\normalfont\small $^{1}$Computer Vision Center, Spain\\
\normalfont\small $^{2}$Universitat Autònoma de Barcelona, Spain\\
\normalfont\small $^{3}$Mohamed bin Zayed University of Artificial Intelligence, United Arab Emirates\\
\normalfont\small $^{4}$Jilin University, China\\
\normalfont\small $^{5}$Program of Computer Science, City University of Hong Kong (Dongguan), China\\[2pt]
\normalfont\small $^{6}$City University of Hong Kong, HK SAR, China\\[6pt]
\normalfont\small Project page: \url{https://tttao-uwu.github.io/DiDE/}
}

\begin{document}

\maketitle

\begin{figure*}[h!]
	\centering
	\vspace{-9mm}
    \includegraphics[width=1\linewidth]{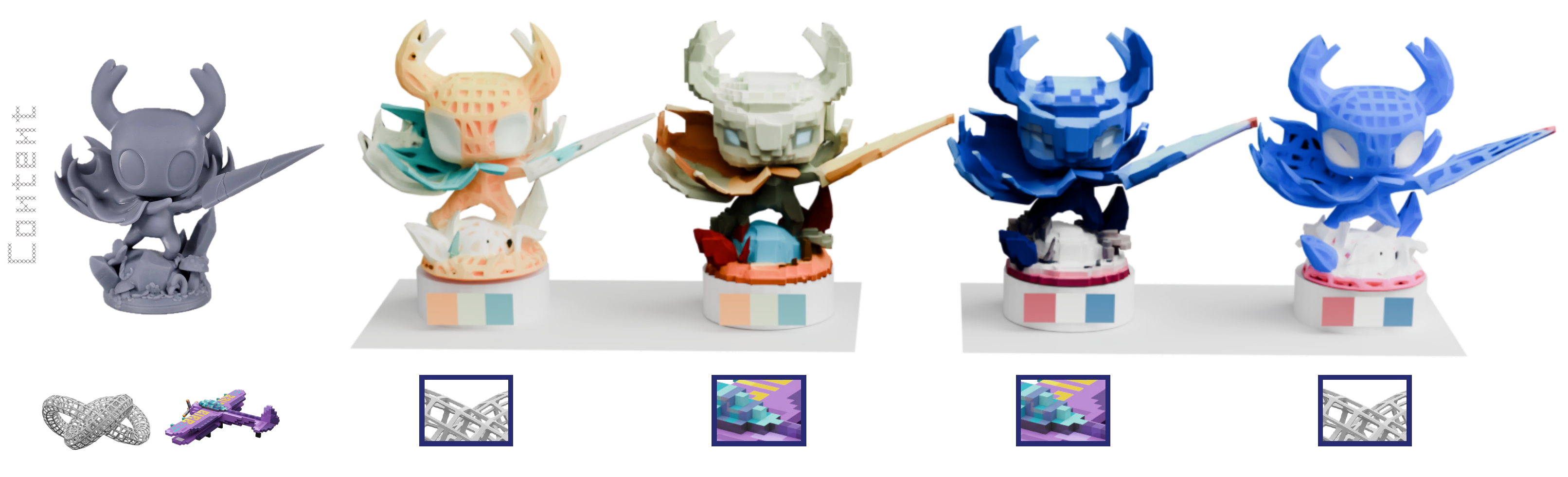}
	\vspace{-7mm}
	\caption{
    Our \textit{training-free} framework \ourmethod enables the generation of stylized 3D assets, with flexible control over the independent modulation of color or texture.
    }
    \label{fig:teaser}
\end{figure*}

\begin{abstract}

Recent advances in rectified flow-based image-to-3D generative models 
have enabled high-fidelity 3D asset generation. Building on this, 
a growing line of work has exploited these strong 3D priors for 
training-free stylization, transferring visual attributes from a 
reference image onto a generated 3D asset. However, existing methods 
enforce an all-or-nothing paradigm: color and texture are transferred 
jointly, with no mechanism to control them independently — a limitation we formalize as \textit{Disentangled 3D Stylization (\ourproblem)}. To address this, we propose \ourmethod, the first training-free framework for \ourproblem. Key to our approach is the observation that the structured latent space of image-to-3D models is overcomplete with respect to texture: texture information occupies only a small subset of the style-significant channels, leaving a free subspace available for independent color encoding. \ourmethod exploits this via a channel partition mechanism that processes a content image, a texture reference, and a color reference through dedicated branches and composes both style signals interference-free at every self-attention layer, preserving content geometry throughout. Experiments on \ourbench, our newly collected multi-reference benchmark, show that \ourmethod consistently outperforms 2D and 3D stylization baselines in color fidelity, texture transfer, and content preservation. 
\end{abstract}

\clearpage

\section{Introduction}
\label{sec:intro}

Generative models for 3D asset creation have witnessed significant advancements ~\cite{yang2024hunyuan3d_v1,hunyuan3d22025tencent_v2,xiang2025trellis2}, facilitating their widespread adoption across diverse domains, including entertainment~\cite{xu2024sketch2scene_3dgame}, robotics~\cite{ke20243d_robot} and healthcare~\cite{khader2023denoising_3dmedical}.
In particular, recent rectified flow-based approaches~\cite{lipman2022flow_matching,liu2022_rectified_flow,hunyuan3d2025hunyuan3d_v21}, exemplified by TRELLIS~\cite{xiang2025trellis2,xiang2025structured_trellis}, have demonstrated the ability to generate high-fidelity 3D objects by jointly modeling sparse structural representations and appearance features within a unified latent space, while explicitly disentangling geometric structures from appearance attributes.
Beyond 3D asset generation, a growing line of work has extended generative models to support \textit{3D stylization}~\cite{hu2026pegasus3d,qu2025stylesculptor,sun2026morphany3d} — the task of transferring stylistic attributes from a reference image onto a generated 3D asset. By conditioning generation on a style reference, these methods allow the produced asset to inherit visual properties such as surface material, color palette, and structural texture, while remaining semantically aligned with a content image. 
This capability carries significant practical value: a single base geometry can be re-stylized for different artistic contexts, production pipelines, or brand identities without requiring manual re-authoring. Although existing stylized 3D generation methods~\cite{qu2025stylesculptor,sun2026morphany3d} have achieved promising results, they still exhibit a critical limitation: style attributes (such as texture and color) cannot be controlled independently, which severely restricts artistic freedom in practice.

In practice, content creators rarely want to adopt a single reference style. Instead, they often wish to selectively borrow specific components while preserving the core of their own artistic vision~\cite{epstein2023art,hertzmann2022toward}. Color palettes play an essential role in this process: they are carefully designed to convey specific emotions or maintain brand consistency, and a creator may wish to preserve their own color scheme while borrowing surface texture from a reference. Equally, one might want to adopt the rough, weathered texture of aged stone from one reference while keeping the warm tones of an entirely different palette. These two attributes — \textit{color and texture} — are visually and semantically distinct, yet current 3D stylization approaches~\cite{,hu2026pegasus3d,qu2025stylesculptor} enforce an all-or-nothing paradigm: applying a reference style requires accepting all its visual attributes simultaneously, which severely restricts artistic freedom and practical applicability.
To the best of our knowledge, only one prior work addresses disentangled stylization in 2D~\cite{qin2025free_sadis}, yet it cannot be generalized to 3D, as style embeddings from separate references cannot simply be composed in 3D latent space. This leaves the problem of fine-grained, decoupled 3D stylization  entirely open.


To address these limitations, we introduce \textit{Disentangled 3D Stylization} (\ourproblem), which aims to decompose reference styles into independently controllable color and texture attributes, enabling artists to selectively transfer specific style elements while preserving content geometry. 
We further propose Direct Injection with color-texture DEcoupling (\ourmethod), the first training-free framework for this task, which takes three inputs — a content image, a texture reference, and a color reference — processed through dedicated branches. To achieve this, We build on prior findings~\cite{qu2025stylesculptor} showing that feature channels of image-to-3D models~\cite{xiang2025structured_trellis} can be decomposed into style- and content-significant subsets, with one branch dedicated to style and another to content. However, directly extending this framework with an additional branch for color is non-trivial: combining two style branches through the same attention pathway leads to feature interference, making disentangled control of texture and color difficult.
To address this problem, we further show that texture information occupies only a subset of the style-significant channels, leaving additional feature capacity available for encoding an independent color attribute. We allocate those informative channels from the texture branch, while populating the remaining free channels with color features from a separate color reference branch. Finally, this partition is applied at every self-attention layer of the texture generation stage, enabling composable, interference-free injection of texture and color while preserving full 3D consistency.


Since no existing method addresses the disentangled stylization problem directly in 3D, we evaluate against complementary baselines on our proposed \ourbench benchmark: SADis~\cite{qin2025free_sadis}, the only prior work on color-texture disentanglement but operating in 2D, and StyleSculptor~\cite{qu2025stylesculptor} and MorphAny3D~\cite{sun2026morphany3d}, which perform 3D stylization but without attribute disentanglement. In summary, our contributions are as follows:
\begin{itemize}[leftmargin=10mm]
    \item We introduce \textit{Disentangled 3D Stylization} (\ourproblem), 
    a novel problem formulation that decomposes reference styles into 
    independently controllable color and texture attributes, which is neglected in prior works.
    
    \item We are the \textit{first} to identify and exploit the overcomplete 
    nature of the structured latent space in image-to-3D models, demonstrating that texture 
    information saturates only a small subset of channels, leaving a free 
    subspace available for independent color and texture injection.
    
    \item We present Direct Inject with color-texture DEcoupling (\ourmethod), the first tuning-free \textit{color-texture 
    disentanglement} framework for 3D stylization, which exploits this 
    overcomplete structure via a channel partition mechanism to enable 
    composable, interference-free style control while preserving content 
    geometry.
    
    \item Experimental results demonstrate that \ourmethod consistently outperforms existing 2D and 3D stylization baselines in color fidelity, texture transfer, and content preservation 
    on our new collected multi-reference disentangled 3D stylization benchmark \ourbench.
\end{itemize}




\section{Related Work}
\label{sec:related_work}

\minisection{3D Generative Models.}
Two dominant paradigms have been established for high-quality 3D asset generation.
The first focuses on distilling useful priors from pretrained 2D generative models into feed-forward 3D reconstruction algorithms.
They either fine-tune 2D models to synthesize multi-view images~\cite{shi2023mvdream,shriram2024realmdreamer,yu2024viewcrafter}, which are further reconstructed into 3D assets via techniques such as 3D Gaussian splatting~\cite{kerbl2023_3dgs};
or directly guides 3D optimization through gradient signals from 2D diffusion models~\cite{poole2023dreamfusion}.
Despite their effectiveness, they lack an explicit latent 3D space to achieve fine-grained structural control.
To overcome this limitation, the second paradigm pursues unified native 3D generative models trained from scratch.
Inspired by the remarkable success of diffusion models in visual synthesis~\cite{ho2020ddpm,khachatryan2023text2video,Rombach_2022_CVPR_stablediffusion}, recent methods~\cite{cheng2023sdfusion,ren2024xcube,vahdat2022lion,zeng2024paint3d} learn compact latent representations and perform generation within this compressed latent space, significantly advancing 3D generative modeling.
Benefiting from large-scale 3D datasets~\cite{deitke2023objaverse_plus,deitke2023objaverse}, modern 3D foundation models~\cite{chen20253dtopia,hong2023lrm,li2025triposg,hunyuan3d2025hunyuan3d_v21} have been developed to capture strong geometric priors, enabling not only high-quality generation across diverse object categories but also various downstream applications~\cite{du2025hierarchical,hu2024neural}.
However, most existing native 3D generators are restricted to either explicit representations 
or implicit formulations.
TRELLIS~\cite{xiang2025structured_trellis} addresses this constraint by introducing a Structured Latent Representation (SLAT) that supports flexible multi-format 3D generation, whose versatility we inherit to accommodate diverse 3D modalities in our framework.
Although these 3D foundation models exhibit strong generalization abilities, they mainly generate samples from a learned data distribution and lack dedicated mechanisms for
example-driven control.
In this work, we leverage the powerful priors from the 3D foundation model TRELLIS~\cite{xiang2025structured_trellis} and introduce a decoupled personalization framework.
Our approach enables flexible adaptation of both geometry and appearance, while fully retaining the high-quality generative capacity of the underlying models.

\minisection{Controllable 3D Generation.}
Recent methods have been developed to enable image-guided control for 3D generation~\cite{dong2024coin3d,gao2025charactershot,huang2024dreamcontrol,wang2024themestation,xu2023dream3d}.
Among these, some approaches inject explicit shape priors to regulate the geometric attributes of synthesized 3D content.
In parallel, another line of work unifies the stylistic appearance of generated objects by leveraging reference images~\cite{richardson2023texture,song2024style3d,xie2024styletex,yeh2024texturedreamer}.
However, such style-guided 3D generation pipelines typically require users to provide predefined 3D geometry such as meshes, and enforce fixed geometric structures throughout generation to maintain spatial consistency.
This strong restriction confines their controllability to texture style only, while leaving geometric style—another critical dimension of 3D asset customization—unchanged and fully uncontrollable.
In this work, we employ TRELLIS as our backbone model, taking advantage of its expressive 3D latent representations that comprehensively encode both geometric structure and textural appearance.
Built on these latent features, we achieve zero-shot style-guided 3D generation that supports simultaneous control over both texture and geometric style.
    
\minisection{Image-based 3D Stylization.}
Given a style reference input and a target content shape, 3D stylization is defined as the task of distilling stylistic patterns from the reference and transferring these patterns to the target geometry to synthesize stylistically aligned 3D assets. This problem has been investigated across diverse 3D representation frameworks, including explicit meshes~\cite{hertz2020deep,hollein2022stylemesh,hu2024neural}, neural implicit fields, and Gaussian splatting~\cite{huang2025edit360,huang2022stylizednerf}.
To enhance generalizability, a prominent line of recent work~\cite{chen2025art_deco,dong2024coin3d,xie2024styletex} harnesses powerful 2D priors from large-scale image diffusion models~\cite{Rombach_2022_CVPR_stablediffusion} for 3D stylization. Concurrently, StyleSculptor~\cite{qu2025stylesculptor} leverages strong 3D priors~\cite{xiang2025structured_trellis} to achieve view-consistent stylization. 
However, all these approaches share a critical limitation: contemporary 3D stylization methods~\cite{hu2026pegasus3d,qu2025stylesculptor} enforce an all-or-nothing paradigm, where adopting a reference style necessitates the simultaneous incorporation of all its visual attributes. This inherent constraint severely limits 
the methods’ practical utility in real-world content creation scenarios.

\section{Methods}
\label{sec:methods}
To address the \ourproblem problem, we propose a disentangled 3D style transfer framework \ourmethod built upon the TRELLIS backbone.  
Given a content image $I_{cnt}$, a texture reference $I_{txu}$, and a color reference $I_{col}$, our goal is to generate a 3D asset that (i) preserves the geometric structure of $I_{cnt}$, (ii) inherits texture from $I_{txu}$, and (iii) adopts color from $I_{col}$, while maintaining cross-view consistency. Unlike conventional 2D style transfer, our approach performs disentangled style modeling in the 3D latent space and injects style during rectified flow-based refinement.

\begin{figure}[!t]
    \centering
    \begin{minipage}{0.54\textwidth}
        \centering
        \includegraphics[width=\linewidth]{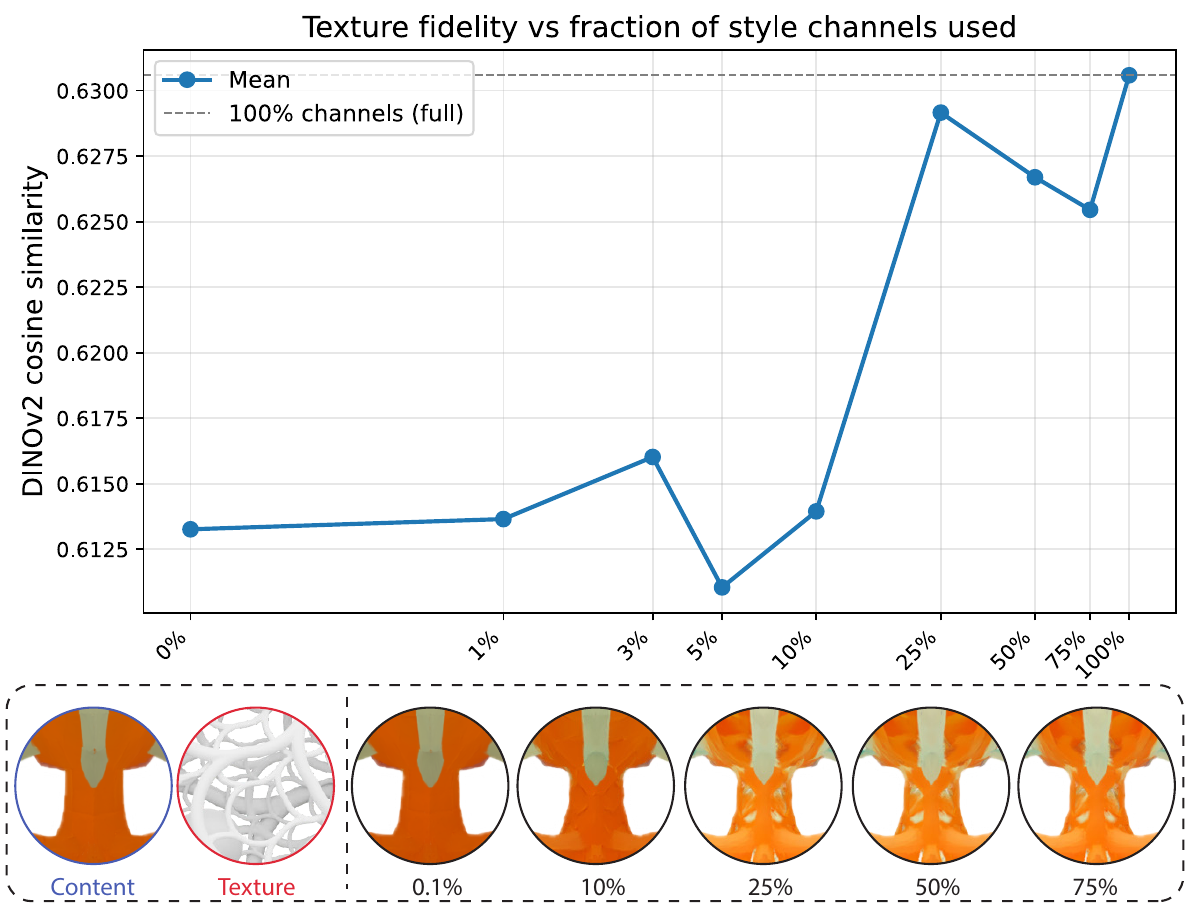}
        \caption{\textbf{Top:} DINOv2 texture similarity rises sharply up to 25\% of channels retained, then plateaus — indicating that texture information is concentrated in a small channel subset. \textbf{Bottom:} Visual inspection shows that texture is reproduced beyond this threshold, with negligible improvement from additional channels.}
        \label{fig:k_analy_channel}
    \end{minipage}
    \hfill
    \begin{minipage}{0.43\textwidth}
        \centering
        \includegraphics[height=0.90\linewidth]{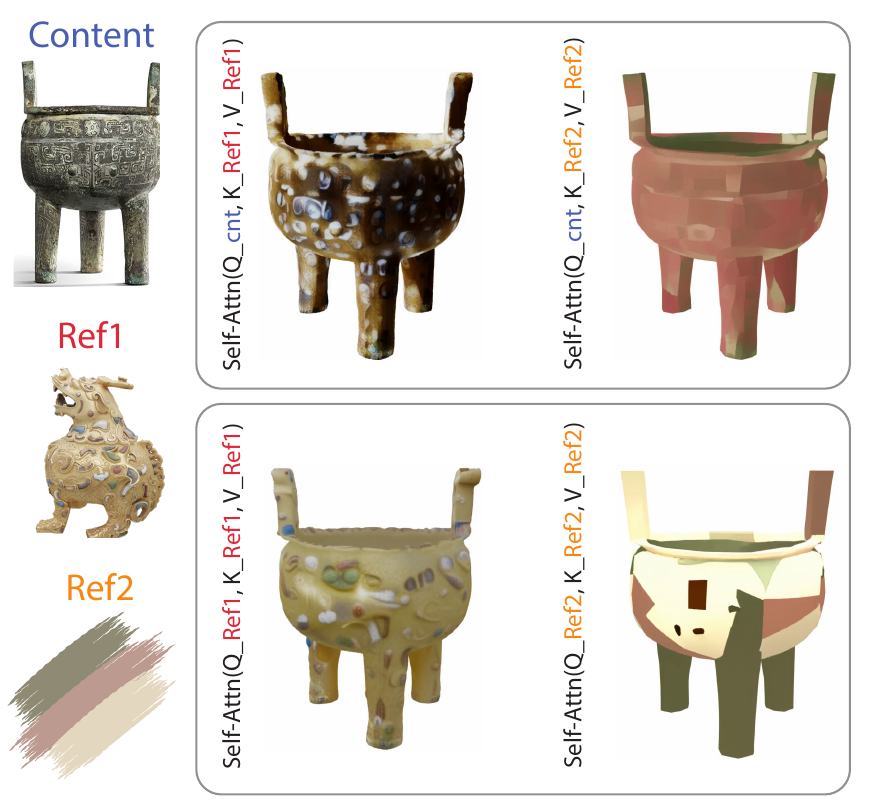}
        \vspace{0.0mm}
        \caption{
        Conventionally, the query is derived from the content image, with the key/value from the reference image. This leads to content bias. In contrast, using query/key/value all from the reference image improves style consistency.}
        \label{fig:analy_query}
    \end{minipage}
\vspace{-3mm}
\end{figure}

\subsection{Preliminaries}
\minisection{TRELLIS}~\cite{xiang2025structured_trellis} is a foundation model for conditional 3D asset generation that synthesizes assets in multiple representations (e.g., NeRF, 3DGS, mesh) from text or image inputs.  
It adopts a two-stage rectified flow framework to progressively transform noise into structured 3D representations: Stage I predicts a sparse geometric structure, and Stage II refines per-voxel latent features for texture and material details.

Following this pipeline, we keep the first-stage sparse structure prediction unchanged to preserve geometric fidelity. Given an input image $I_{cnt}$, we use the DINOv2 model~\cite{oquab2023dinov2} to encode it into a latent $z_c = \mathcal{E}(I_{cnt})$. The Voxel Flow Transformer $\mathcal{VFT}$ then predicts a sparse voxel set
\begin{equation}
P = \{p_i\}_{i=1}^{L}, \quad p_i \in \{0,1,\ldots,N-1\}^3
\end{equation}
where $p_i$ is the positional index of an active voxel in the 3D grid intersecting with the surface of object, $L$ is the total number of active voxels, and $N$ is the maximum voxel index per dimension.
We feed the coarse location information obtained in the first stage, along with the image embedding, into Stage II to generate features for each voxel. In the second stage, we complete the style transfer.


\subsection{Overcompleteness of the TRELLIS Feature Space}\label{sec:overcompleteness}

StyleSculptor~\cite{qu2025stylesculptor} showed that TRELLIS feature channels can be decomposed into style and content-relevant subsets via patch-wise variance. However, their formulation assumes a single style reference. In our setting, texture and color must be injected independently into the same style branch, which can lead to feature interference. We therefore investigate whether the style-significant channels are overcomplete, asking whether all of them are actually required to faithfully encode texture. We conduct a channel ablation experiment in which we progressively reduce the number of channels used for texture injection at every DiT block of TRELLIS stage II, retaining only the top-k highest-variance channels from the style branch of Stage II. For each value of k, we generate outputs across the full set of style references from StyleSculptor~\cite{qu2025stylesculptor} using a fixed content image, render a single view, and measure DINOv2 cosine similarity between the mean patch features of the rendered output and the corresponding style reference image. 
As shown in Fig.~\ref{fig:k_analy_channel}, DINOv2 texture similarity is $\approx 0.6125$ when only a small fraction of channels is retained, rises to $\approx 0.625$--$0.63$ once more than $25\%$ of the channels are retained, and remains within this range up to $100\%$.
To further support this finding, we conduct a PCA reconstruction analysis (see appendix), confirming the low intrinsic dimensionality of texture within the TRELLIS feature space, and validating that style-significant channels contain a large free subspace that can be independently allocated to a second style attribute.


\subsection{Color-Texture Attributes Embedding}


Given a content image $I_{cnt}$, a texture reference $I_{txu}$, and a color reference $I_{col}$, our framework processes each input through a dedicated branch. The content branch is kept unchanged from the standard image-to-3D pipeline~\cite{xiang2025structured_trellis} to preserve geometric structure, while the texture and color branches apply dedicated preprocessing to extract disentangled style embeddings, as described below.

\minisection{Texture Reference.}
We convert the texture reference image $I_{txu}$
into grayscale to suppress color information, yielding $I_{txu}^{\text{gray}}$. This grayscale image is then fed into the DINOv2~\cite{oquab2023dinov2} encoder to obtain the texture embedding $f_{txu} = \mathcal{E}\big(I_{txu}^{\text{gray}}\big)$.

\minisection{Color Reference.}
Our method supports two types of color reference inputs: \textit{reference images}, common in style transfer applications, and \textit{color palettes}, widely used in multimedia and design workflows. Both are processed to isolate color information from texture structure before encoding.

When a reference image is used as the color reference, we extract its low-frequency components via a Fourier transform while suppressing high-frequency signals. Intuitively, high-frequency components primarily encode textural details (e.g., edges and fine patterns), whereas low-frequency components capture global color distributions. By filtering out high-frequency signals, our method enforces a structure-invariant color conditioning, thereby improving the disentanglement between color and texture.

Given a color palette $\{ \mathbf{c}_k \}_{k=1}^{n_{\mathrm{colors}}}$, we construct a spatially varied image $I_{\mathrm{vor}}$ via Voronoi tessellation: we sample random seeds over the image plane, assign each a palette color, and map every pixel $\mathbf{x}$ to its nearest seed:
\begin{equation}
    I_{\mathrm{vor}}(\mathbf{x}) = \mathbf{c}_{a_{j^*(\mathbf{x})}}, \quad j^*(\mathbf{x}) = \arg\min_j \| \mathbf{x} - \mathbf{s}_j \|_2^2.
\end{equation}
This produces an image that encodes the palette's color distribution without any texture structure.
In both cases, the processed image is encoded via DINOv2~\cite{oquab2023dinov2} to obtain the color embedding $f_{\mathrm{col}} = \mathcal{E}(I_{\mathrm{vor}})$.

\subsection{Disentangled 3D Style Transfer (\ourproblem)}

\begin{figure}
    \centering
    \includegraphics[width=1\linewidth]{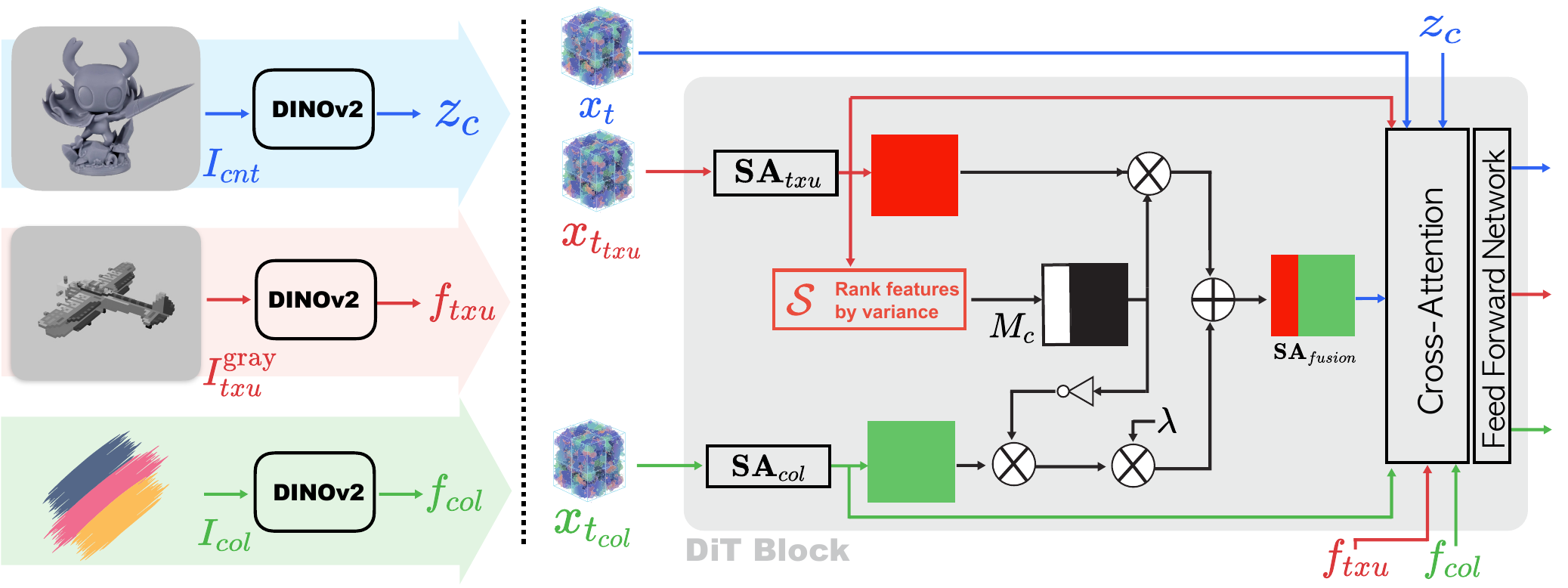}
    \vspace{-5mm}
    \caption{\textbf{Overview of \ourmethod.} 
\textit{Left:} Our framework takes three inputs — a content image, 
a texture reference, and a color reference — processed through three 
independent branches: the content branch ({\color{blue}blue}), the 
texture branch ({\color{red}red}), and the color branch 
({\color{green}green}).
\textit{Right:} At each DiT block, the texture and color branches 
produce independent self-attention outputs. A channel selection 
operator $\mathcal{S}$ partitions the feature channels into 
texture-allocated (high-variance) and color-allocated (low-variance) 
subsets, and the two attention outputs are composed via the binary 
mask $\mathbf{M_c}$ into a single fused representation 
$\mathbf{SA}_{\text{fusion}}$, which is then injected into the 
residual stream of the content branch through its gated update.}
    \vspace{-4mm}
    \label{fig:method}
\end{figure}

\ourmethod runs three parallel branches through the transformer blocks 
of Stage II — one for the content image, one for the texture 
reference, and one for the color reference — each processing its input 
through independent self-attention layers. At every transformer block, 
the texture and color branches produce separate attention outputs, which 
are composed via a channel partition mechanism into a single fused 
representation. This fused output is then injected into the content 
branch, overwriting its self-attention output before the residual update. 
We describe each step in detail below.

\minisection{Per-Branch Self-Attention.}
Given an input feature sequence $\mathbf{X} \in \mathbb{R}^{N \times d}$,
a standard self-attention layer computes as:
\begin{equation}
\mathrm{Attn}(\mathbf{X}) = \mathrm{Softmax}\!\left(
\frac{\mathbf{Q}\mathbf{K}^{\top}}{\sqrt{d}}
\right)\mathbf{V}, \quad
\mathbf{Q} = \mathbf{X}W_Q,\;
\mathbf{K} = \mathbf{X}W_K,\;
\mathbf{V} = \mathbf{X}W_V.
\end{equation}

Prior 3D stylization methods~\cite{qu2025stylesculptor} adopt a 
cross-attention formulation in which the query is derived from the 
content image while keys and values come from the style reference. 
As shown in Fig.~\ref{fig:analy_query}, this leads to structural 
leakage: the content query biases the output toward the content 
geometry and weakens style transfer.

We instead run each style branch with its own standard self-attention, 
using query, key, and value all derived from the same branch input:
\begin{align}
\mathbf{SA}_{txu} &= \mathrm{Attn}(\mathbf{Q}_{txu}, \mathbf{K}_{txu}, \mathbf{V}_{txu}), \\
\mathbf{SA}_{col} &= \mathrm{Attn}(\mathbf{Q}_{col}, \mathbf{K}_{col}, \mathbf{V}_{col}).
\end{align}
This ensures that each branch faithfully encodes the style of its 
own reference without contamination from the content image.

\minisection{Texture-Color Composition.}
Building on our finding that texture information saturates only a 
subset of the style-significant channels (Sec.~\ref{sec:overcompleteness}), 
we exploit the remaining free channels to encode an independent color 
attribute. This channel-wise decomposition is applied at every 
transformer block of Stage II, as illustrated in 
Fig.~\ref{fig:method} (right).

We introduce a channel selection operator $\mathcal{S}$, where 
$\mathcal{S}(h) \subset \{1, \dots, C\}$ denotes the indices of the 
top $k\%$ highest-variance channels of a feature map with $C$ channels. 
These channels are allocated to the texture branch, while the 
complementary set is allocated to the color branch. We construct 
a binary mask $\mathbf{M} \in \{0,1\}^C$:
\begin{equation}
M_c =
\begin{cases}
1, & c \in \mathcal{S}(h), \\
0, & \text{otherwise,}
\end{cases}
\end{equation}
and define the fused attention output as:
\begin{align}
\mathbf{SA}_{\text{fusion}} = \mathrm{Attn}\!\left(\mathbf{Q}_{txu},
\mathbf{K}_{txu}, \mathbf{M} \odot \mathbf{V}_{txu}\right)
+ \lambda\, \mathrm{Attn}\!\left(\mathbf{Q}_{col}, \mathbf{K}_{col},
(1 - \mathbf{M}) \odot \mathbf{V}_{col}\right),
\label{eq:sa_fusion}
\end{align}
where $\odot$ denotes element-wise multiplication broadcast along 
the sequence dimension, and $\lambda > 0$ is a scalar hyperparameter 
that controls the trade-off between the color and texture branch.

\minisection{Injection into the Content Branch.}
The fused representation $\mathbf{SA}_{\text{fusion}}$ is injected 
into the content branch by replacing its self-attention output at 
each transformer block. Following the standard gated residual 
mechanism of the DiT blocks in TRELLIS, the content features are 
updated as:
\begin{equation}
\mathbf{X}_{cnt}' = \mathbf{X}_{cnt} + \alpha_{cnt} \odot \mathbf{SA}_{\text{fusion}},
\end{equation}
where $\alpha_{cnt} \in \mathbb{R}^{d}$ is the learned per-channel 
gate derived from the content branch's adaptive layer norm 
conditioning, following the AdaLN-zero parameterization~\cite{Peebles2022DiT}. 
This preserves the spatial structure and diffusion conditioning of the content branch while fully transferring the composed style signal, keeping the generated geometry faithful to the content image throughout the generation.

\begin{figure}[t]
    \centering
    \includegraphics[width=0.999\textwidth]{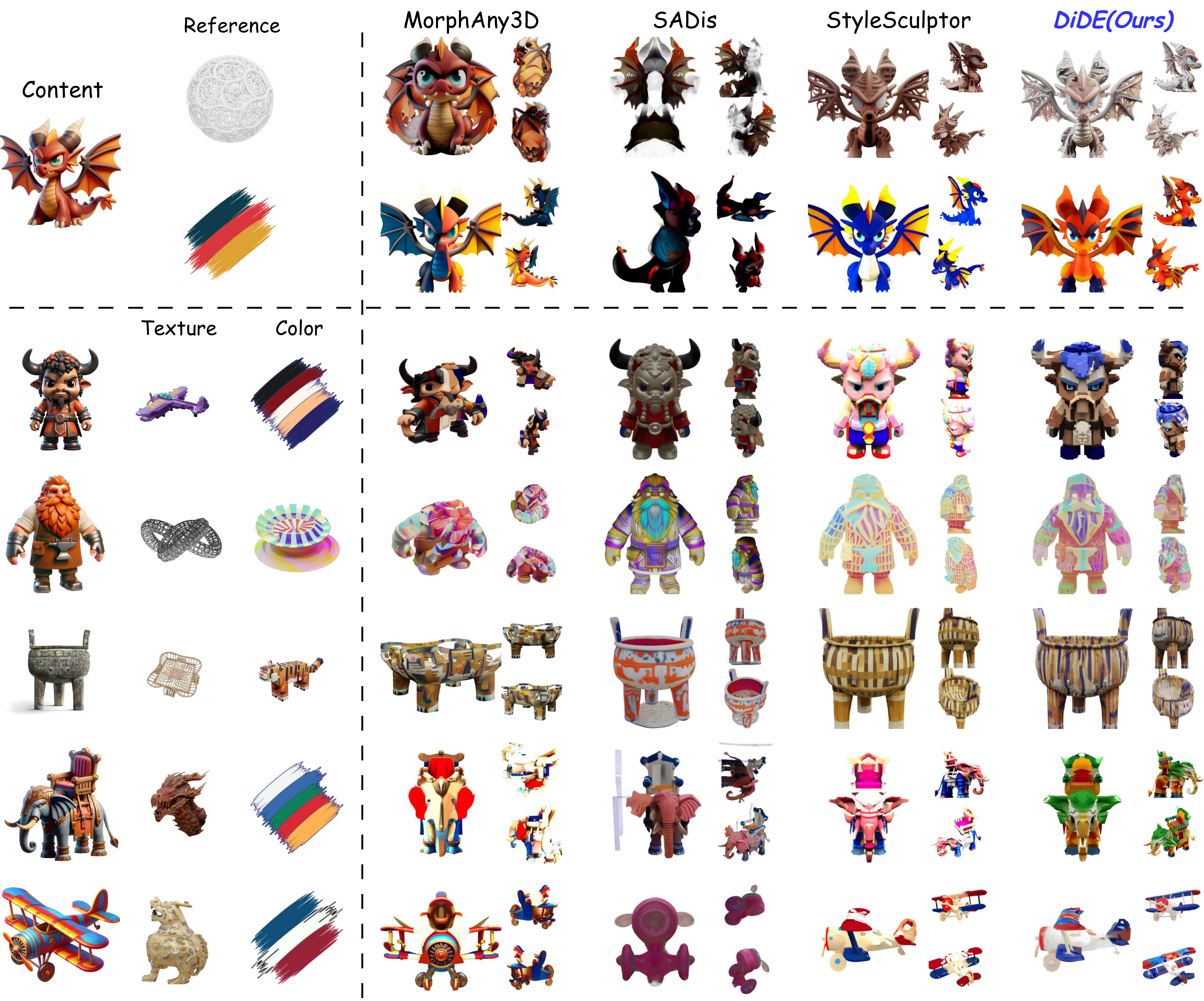}
    \vspace{-5mm}
    \caption{Comparison with existing 3D stylization methods and adapted baselines for the \ourproblem problem. The first two rows show results with a single style reference, while the remaining rows show results with disentangled texture and color references.}
    \vspace{-3mm}
    \label{fig:compares}
\end{figure}

\begin{figure}[t]
    \centering
    \includegraphics[width=0.999\textwidth]{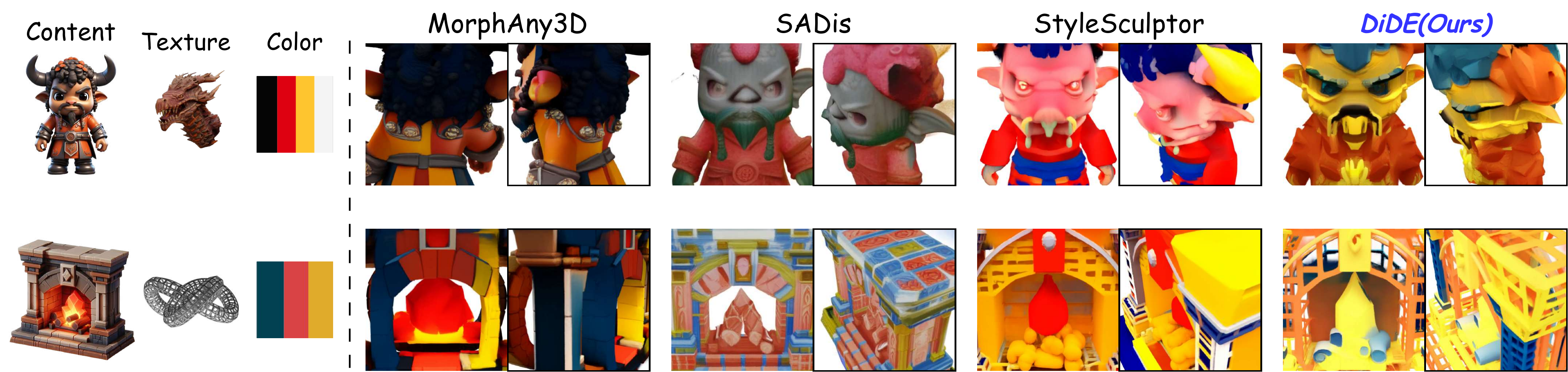}
    \vspace{-5mm}
    \caption{Zoomed-in comparison of the results. The enlarged views provide a closer inspection of geometry preservation and the disentangled transfer of texture and color.}
    \vspace{-3mm}
    \label{fig:compares_enlarge}
    \vspace{-3mm}
\end{figure}

\section{Experiments}
\label{sec:experiments}

\subsection{Experimental Setup}

We conduct extensive experiments to validate three core claims of \ourmethod: (i) effective disentanglement of color and texture in the generated 3D assets, (ii) faithful preservation of the content geometry, and (iii) competitive style fidelity against existing methods. Since no prior benchmark addresses the \ourproblem setting, we introduce \ourbench and compare against representative 2D and 3D stylization baselines.

\minisection{Datasets.}
We collected 37 images to form a novel multi-image style transfer dataset named \ourbench for the \ourproblem setup. These images include 10 content images, 12 texture reference images, and 15 palette images. The content images and texture reference images were selected from Stylesculptor~\cite{qu2025stylesculptor} and Morph3D~\cite{sun2026morphany3d}, and each palette image contains 3-5 different colors. These image combinations generate 1,800 3D assets for subsequent evaluation.

\minisection{Implementation Details.}
We mainly build our method \ourmethod upon TRELLIS~\cite{xiang2025structured_trellis} while also verified on TRELLIS.2~\cite{xiang2025trellis2}.
For the hyperparameters, we set $k=25, \lambda = 1.2$.
Both stages use a CFG strength of 3.0, with 25 inference steps.
The reproducing code will be released after the peer review process.
All experiments are conducted on a NVIDIA A40 GPU.

\minisection{Baseline Methods.}
We compare against three representative training-free stylization approaches, 
each adapted to the \ourproblem two-reference setting to the best extent their 
architectures permit. SADis~\cite{qin2025free_sadis} is the only prior method 
addressing color-texture disentanglement, though in 2D: we apply it to the 
content image and lift the stylized result to 3D using TRELLIS. 
MorphAny3D~\cite{sun2026morphany3d} natively supports two image inputs via its 
dual-target morphing setting, which we use to provide the texture and color 
references directly. StyleSculptor~\cite{qu2025stylesculptor}, originally 
designed for a single content-texture pair, is adapted by adding an extra 
reference branch for the color reference; the self-attention outputs of the 
texture and color branches are then summed to produce the fused style 
representation. We emphasize that StyleSculptor and MorphAny3D were not designed for disentangled two-reference stylization; these adaptations represent the closest reasonable comparison points rather than intended use cases. SADis, while designed for color-texture disentanglement, operates in 2D and requires an 
additional lifting step to produce 3D assets. Additional 
details on the experimental setup are provided in the appendix.


\minisection{Evaluation Metrics.}
We adopt several quantitative metrics across four axes: (1) For \textit{Geometry  
Semantics}, CLIP-I~\cite{radford2021clip} measures semantic similarity 
between the generated asset and the content image; and Chamfer Distance and Hausdorff Distance are used to measure the geometric difference between two assets. (2) For \textit{Texture 
preservation}:  LPIPS~\cite{zhang2018lpips} 
captures perceptual distortion and potential style leakage; and 
SSIM~\cite{wang2003ssim} evaluates structural similarity.  (3) For \textit{color 
fidelity}: MS-SWD~\cite{he2024_MS_SWD} measures color distance between the 
generated asset and the color reference, and C-Hist computes the distance between color histograms. 
(4) Finally, we employ 
\textit{VLM}-based evaluation using InternVL2.5-8B~\cite{chen2024internvl25}, 
which performs binary judgments on texture consistency (whether the generated 
asset shares structural style with the texture reference, ignoring color) and 
color fidelity (whether colors absent from the color reference are introduced). 
Full prompts and scoring protocol are provided in 
Appendix~\ref{appendix:vlm_eval}. All metrics are computed over 1,800 generated assets.

\subsection{Experimental Results}

\minisection{Qualitative Results.}
In our evaluation of generation quality, visual results are presented in Fig.~\ref{fig:compares} and Fig.~\ref{fig:compares_enlarge}. 
For SADis, the method exhibits subpar object detail preservation, while introducing redundant textures and color artifacts in the background region. After 3D rendering, the reconstructed textures and colors also fail to align closely with the reference images.
For StyleSculptor, the model fails to achieve full disentanglement of texture and color. The generated 3D assets exhibit a mixed color distribution derived from both reference images; furthermore, the results often suffer from style bias, tending to favor either the texture reference or the color reference rather than achieving balanced, independent control over both attributes.
Regarding MorphAny3D, its Morphing Cross-Attention mechanism causes the colors and styles inherent in the content image to exert a significant influence on the final generation result. Additionally, the method generates spurious colors that are not present in the provided color palette images.
In contrast, our \ourmethod enables precise decoupling of texture structure and color attributes. For instance, we can accurately align the cutout structure of the texture reference with the target color— a capability not achievable by other baseline methods. Simultaneously, our method \ourmethod ensures that the generated 3D asset remains consistent with the content image, free from distortions, flips, or other visual artifacts.

\begin{table}[t]
\centering
\caption{Quantitative comparisons with existing style transfer methods. 
The best and second-best results are highlighted in \textbf{bold} and \underline{underlined}, respectively. 
CD and HD denote Chamfer Distance and Hausdorff Distance, respectively.}
\resizebox{0.99\linewidth}{!}{
\begin{tabular}{c c c c c c c c c c}
\toprule
\multirow{2}{*}{\textbf{Method}} &
\multicolumn{3}{c}{\textbf{Geometry}} &
\multicolumn{3}{c}{\textbf{Color}} &
\multicolumn{3}{c}{\textbf{Texture}}  \\
\cmidrule(lr){2-4}\cmidrule(lr){5-7}\cmidrule(lr){8-10}
&  \textbf{CLIP-I$\uparrow$} & \textbf{CD}$_{^{\times 10^2}}$$\downarrow$ & \textbf{HD}$_{^{\times 10^2}}$$\downarrow$ & \textbf{MS-SWD$\downarrow$} & \textbf{C-Hist$\downarrow$} & \textbf{VLM$\uparrow$} 
& \textbf{LPIPS$\downarrow$}  & \textbf{SSIM}$_{^{\times 10^3}}$$\uparrow$ & \textbf{VLM}$\uparrow$ \\
\midrule
SADis~\cite{qin2025free_sadis}           & 0.73 &18.17 & 41.59 &  \underline{18.38} & 0.99 & 37.17 & \textbf{0.65}  & 116.10  & 18.00 \\
StyleSculptor~\cite{qu2025stylesculptor} & \underline{0.74} & \underline{17.91} & \underline{8.38} &  21.26 & \underline{0.89} & 35.44 & 0.69  & \underline{124.61}  & \underline{23.22} \\
MorphAny3D~\cite{sun2026morphany3d}      & 0.71 & 198.52 &44.50 & \textbf{18.31} & 0.91 & \underline{40.05} & \textbf{0.65}  & 117.21  & 19.72 \\    
\ourmethod (Ours)                        & \textbf{0.75} & \textbf{15.38} & \textbf{7.87} & 20.40 & \textbf{0.87} & \textbf{70.94} & \underline{0.68}  & \textbf{125.19}  & \textbf{30.72} \\
\bottomrule
\end{tabular}
}
\label{tab:quanti_comparison}
\vspace{-4mm}
\end{table}


\minisection{Quantitative Results.}
Table~\ref{tab:quanti_comparison} reports the full results. \ourmethod achieves 
the best CLIP-I score (0.75), indicating superior semantic alignment with the 
content image. For Geometry, \ourmethod achieves the lowest Chamfer (15.38) and Hausdorff (7.87) distances, improving over StyleSculptor by approximately 14.1\% and 6.1\%, respectively. Although SADis obtains a comparable Chamfer distance, its substantially higher Hausdorff distance reveals larger worst-case local deviations, showing that \ourmethod better preserves both global surface structure and local geometry.
For color fidelity, \ourmethod leads on C-Hist (0.87) and by a 
large margin on VLM color accuracy (70.94 vs.\ 40.05 for the next best method), 
confirming faithful and disentangled color transfer. While MorphAny3D and SADis 
yield slightly better MS-SWD scores (18.31 and 18.38 vs.\ 20.40), this comes at 
the cost of degraded texture structure, as reflected in their lower SSIM scores. 
For texture quality, \ourmethod attains the highest SSIM (125.19) and VLM 
texture accuracy (30.72 vs.\ 23.22 for StyleSculptor), demonstrating superior 
structural preservation and texture fidelity. LPIPS scores are comparable across 
all methods (0.65--0.69), suggesting similar levels of perceptual distortion 
relative to the content image.

\subsection{Ablation Study}

\begin{figure}[t]
    \centering
    \includegraphics[width=0.999\textwidth]{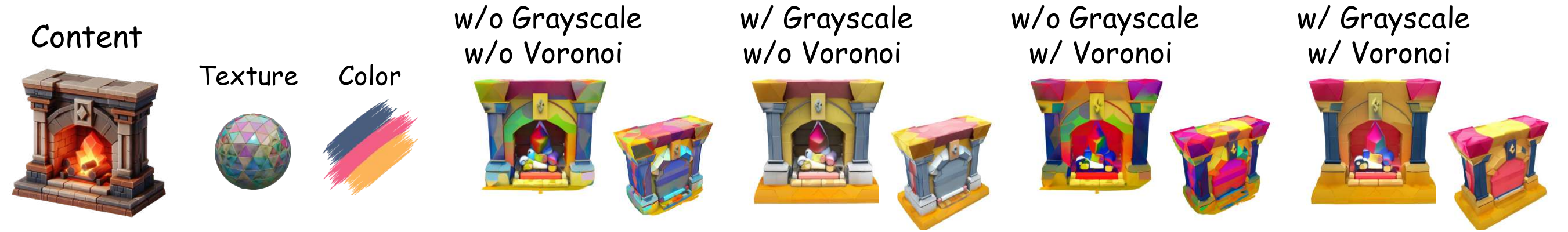}
    \vspace{-5mm}
    \caption{Ablation on grayscale conversion and Voronoi tessellation preprocessing.}
    \vspace{-2mm}
    \label{fig:ablation_comp}
\end{figure}

\minisection{Effect of Branch Preprocessing.}
We ablate the two key preprocessing components of \ourmethod — grayscale conversion of the texture reference and Voronoi tessellation of the color palette — by removing each in turn while keeping the rest of the pipeline unchanged. As shown in Fig.~\ref{fig:ablation_comp}, removing grayscale conversion leaves residual color information in the texture branch, introducing undesired color interference in the generated output. Removing Voronoi tessellation causes color leakage, producing colors that deviate from the reference palette.

\begin{figure}[t]
    \centering
    \includegraphics[width=0.999\textwidth]{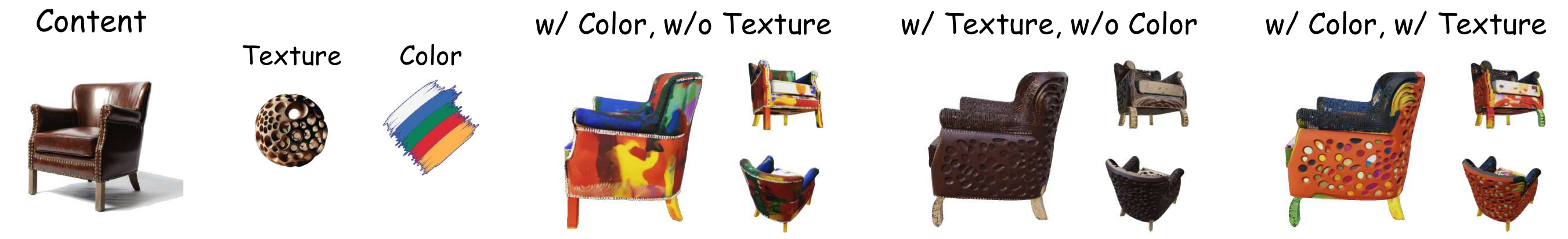}
    \vspace{-5mm}
    \caption{\ourmethod applied to the TRELLIS.2 backbone, showing generalization across model variants.}
    \vspace{-2mm}
    \label{fig:ablation_trellis2}
\end{figure}

\minisection{Model Adaptability.}
To verify the generalizability of \ourmethod beyond the base TRELLIS backbone, we apply it to TRELLIS.2~\cite{xiang2025trellis2}, which extends TRELLIS with a three-stage pipeline where texture and color features are processed in separate stages. As illustrated in Fig.~\ref{fig:ablation_trellis2}, \ourmethod maintains robust and consistent disentanglement performance on TRELLIS.2, confirming that our channel partition mechanism transfers across model variants. The natural stage-wise separation in TRELLIS.2 further aligns with our texture-color decoupling design, making the adaptation straightforward.

\begin{figure}[!t]
    \centering
    \includegraphics[width=0.999\textwidth]{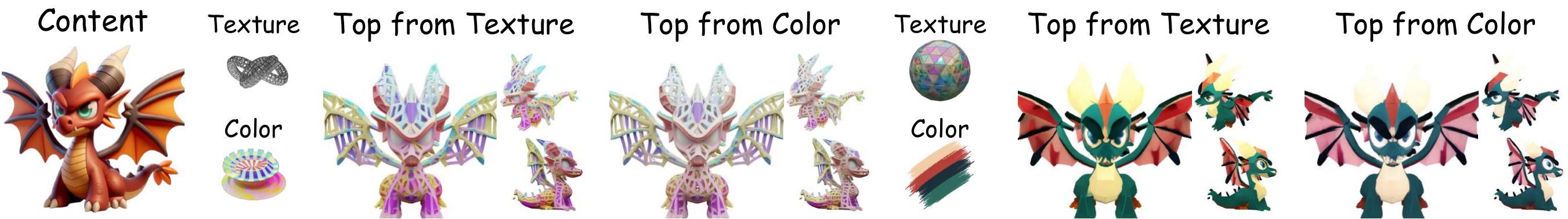}
    \vspace{-5mm}
    \caption{Ablation on channel selection variance source: texture vs.\ color branch.}
    \vspace{-2mm}
    \label{fig:ablation_top_channel}
    \vspace{-3mm}
\end{figure}

\minisection{Channel Selection.}
In \ourmethod, channel variance is computed from the texture branch to determine the channel partition. To assess the sensitivity of this design choice, we ablate by computing variance from the color branch instead, swapping which branch drives the partition. As shown in Fig.~\ref{fig:ablation_top_channel}, both settings achieve consistent texture-color disentanglement, with only minor differences in color saturation. This robustness to the choice of variance source corroborates our earlier finding (Sec.~\ref{sec:overcompleteness}) that the latent feature space is overcomplete, with texture and color information redundantly distributed across channels regardless of which branch is used for estimation. We use texture-branch variance as default since it produces marginally more faithful color rendition in practice.

\begin{wraptable}{r}{0.5\linewidth}
\centering
\caption{Ablation on color branch weight $\lambda$.}
\vspace{-2mm}
\resizebox{0.98\linewidth}{!}{
\begin{tabular}{c c c c c}
\toprule
\multirow{2}{*}{\textbf{$\lambda$}} & 
\multicolumn{2}{c}{\textbf{Color}} &
\multicolumn{2}{c}{\textbf{Texture}}  \\
\cmidrule(lr){2-3}\cmidrule(lr){4-5}
& \textbf{MS-SWD$\downarrow$} & \textbf{C-Hist$\downarrow$} 
& \textbf{LPIPS$\downarrow$}  & \textbf{SSIM}$_{^{\times 10^3}}$$\uparrow$ \\
\midrule
0.5 & 20.24 & 0.92 & 0.68 & 83.36 \\
1.0 & 19.47 & 0.89 & 0.69 & 83.34  \\
\rowcolor{gray!12}
1.2 & 19.25 & 0.89 & 0.69 & 83.45 \\
1.5 & 19.32 & 0.90 & 0.70 & 83.30 \\
2.0 & 19.38 & 0.90 & 0.70 & 83.40  \\
\bottomrule
\end{tabular}
}
\label{tab:ablation_lamda}
\vspace{-10pt}
\end{wraptable}

\minisection{Hyperparameter $\lambda$.}
We ablate the color branch weight $\lambda$ in Eq.~\ref{eq:sa_fusion} by sweeping over the range $[0.5, 2.0]$. All results are averaged over 600 randomly samples. As reported in 
Table~\ref{tab:ablation_lamda}, too small a value of $\lambda$ weakens color transfer fidelity, while too large a value begins to compromise texture quality. This reflects the expected trade-off between the two branches, and 
$\lambda=1.2$ strikes the best balance, achieving the lowest C-Hist (0.89) and MS-SWD (19.25) while maintaining competitive texture scores.

\begin{wraptable}{r}{0.5\linewidth}
\centering
\vspace{-4mm}
\caption{User study results on geometry, texture, and color preference. Higher is better.}
\label{tab:user_study}
\vspace{-2mm}
\resizebox{0.98\linewidth}{!}{%
\begin{tabular}{lccc}
\toprule
Method & Geometry $\uparrow$ & Texture $\uparrow$ & Color $\uparrow$ \\
\midrule
MorphAny3D    & 4.4\%  & 1.2\%  & 2.3\%  \\
SADiS         & 12.1\% & 3.2\%  & 8.7\%  \\
StyleSculptor & 31.8\% & 37.4\% & 16.6\% \\
\textbf{Ours} & \textbf{51.7\%} & \textbf{58.2\%} & \textbf{72.4\%} \\
\bottomrule
\end{tabular}%
}
\vspace{-5pt}
\end{wraptable}

\minisection{User Study.}
Our method received the highest preference rates across all three aspects, with 51.7\% for geometry preservation, 58.2\% for texture consistency, and 72.4\% for color consistency. These results provide complementary perceptual evidence that the quantitative improvements of our method translate into noticeable visual benefits, including better preservation of content geometry and more faithful, disentangled transfer of texture and color. 

\begin{wrapfigure}{r}{0.58\linewidth}
    \centering
    \vspace{-5mm}
    \includegraphics[width=\linewidth]{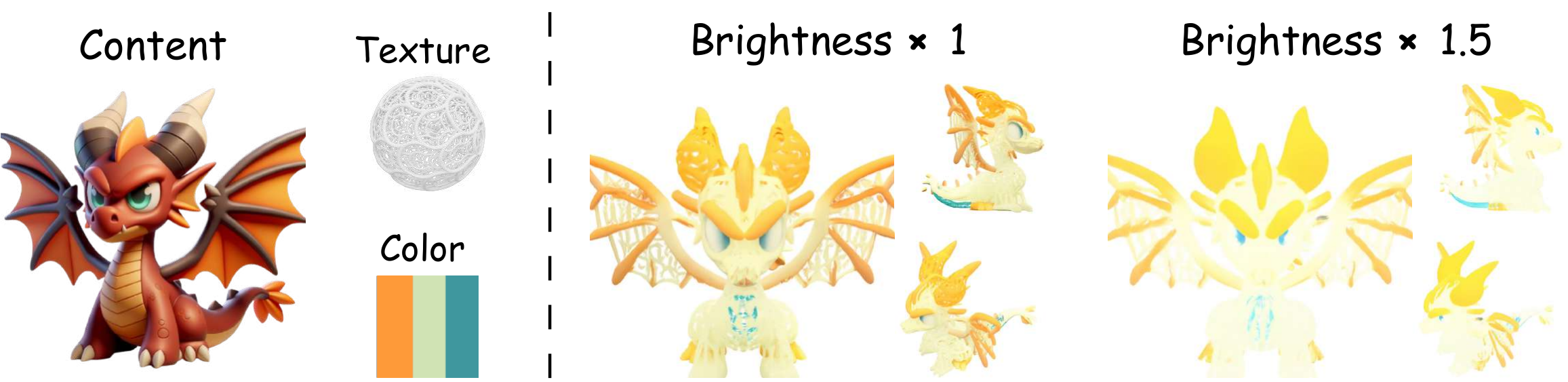}
    \caption{Failure case under texture-reference brightness shifts.}
    \label{fig:failures}
    \vspace{-3mm}
\end{wrapfigure}

\minisection{Failure Cases.}
We further show a failure case in Fig.~\ref{fig:failures}. 
Our method is sensitive to the luminance distribution of the texture reference. 
While the original texture reference produces reasonable texture and color transfer, 
increasing its brightness by $1.5\times$ causes the generated 3D assets to become nearly white, 
with substantial loss of both texture and color details. 
This indicates that large brightness shifts in the reference image remain a limitation of our current appearance-transfer formulation.

\section{Conclusion}
\label{sec:conclusion}


This work addresses a fundamental limitation in existing 3D stylization methods: the inability to independently control distinct stylistic attributes such as color and texture. We formalize this gap as \textit{Disentangled 3D Stylization} (\ourproblem) and propose Direct Injection with color-texture DEcoupling (\ourmethod), the first training-free framework to address it. \ourmethod decouples texture and color by processing each reference through a dedicated branch with tailored preprocessing  and composes them via a channel partition mechanism injected into the 
self-attention layers of a pretrained image-to-3D model, preserving content geometry throughout. Extensive qualitative and quantitative evaluations on \ourbench confirm that \ourmethod outperforms state-of-the-art baselines in disentangled color and texture control. We hope this work opens the door to finer-grained artistic control in 3D content creation and inspires future research on multi-attribute style disentanglement.

\section{Acknowledgments}

Kai Wang acknowledges funding No.R002026G0116 and No.R002026B0121 from Guangdong and Hong Kong Universities 1+1+1 Joint Research Collaboration Scheme, No.TIC2026SP005 from Seed Project under City University of Hong Kong (Dongguan) Technology and Innovation Center, NSFC Grant No. 62606428, funding No.2025QN12X071 from Guangdong Pearl River Talent Program and the start-up grant B01040000108 from CityU-DG. 
Joost van de Weijer acknowledges Grant PID2025-175937NB-I00 and Grant AIA2025-163919-C52, both funded by MICIU/AEI/10.13039/501100011033.
Alexandra Gomez Villa acknowledges the EuroHPC Joint Undertaking for awarding access to the Leonardo supercomputer at CINECA, Italy, under project EHPC-AIF-2026SC01-004.


\clearpage







\clearpage 

\bibliographystyle{plain}
\bibliography{longstrings,mybib}

\clearpage 

\appendix

\section{Statements}
\label{appendix:statements}

\minisection{Limitations.} 
While \ourmethod effectively mitigates detail attenuation and enhances fine-grained texture fidelity for training-free disentangled 3D stylization, it has several limitations: the framework is only validated on rectified flow-matching based native 3D diffusion models (the TRELLIS/TRELLIS.2 family) and has not been tested for generalization to other mainstream 3D generation paradigms, including voxel-based models, signed distance field (SDF) generators, 3D Gaussian splatting diffusion pipelines, and knowledge distillation-based image-to-3D methods.
Furthermore, although the computational overhead is negligible, the low-noise restart refinement step adds mild extra inference operations, restricting direct deployment on extremely memory-constrained edge devices.
Finally, \ourmethod focuses on color-texture decoupling and texture enhancement without a dedicated geometric refinement module, so the geometric quality of the generated 3D assets is inherently dependent on the geometric reconstruction capability of the underlying pre-trained 3D foundation model;
and the current design only supports independent control of color and texture, without extending to multi-attribute disentanglement (e.g., material, lighting, geometric style), which limits its flexibility for complex artistic creation.

\minisection{Broader Impacts.}
\ourmethod advances controllable 3D content creation and delivers widespread practical value by enabling training-free, disentangled color-texture control for 3D stylization, which drastically lowers the technical barrier and production cost for generating high-fidelity and customizable 3D assets, benefiting diverse industries and scenarios including game development, film and television virtual production, industrial design, cultural heritage digitization, robotic perception, and metaverse content construction, empowering both professional creators and non-expert users to flexibly customize 3D assets with independent style control, improving creative efficiency and diversifying content supply; on the other hand, the strong detail-preserving and high-fidelity texture transfer capability of \ourmethod introduces potential intellectual property and ethical risks, as the model could be misused for unauthorized duplication, counterfeiting, or commercial exploitation of copyrighted objects, branded products, real-world artifacts, or cultural heritage items without permission, and to address such risks, we advocate for the responsible deployment of the technology, enforcing content authentication for generated assets, establishing clear usage norms and legal compliance guidelines, and promoting ethical adoption across industries to ensure that \ourmethod creates public value while minimizing harms from misuse.

\minisection{Ethical Statement.}
We acknowledge the potential ethical implications associated with generative image-to-3D technologies, including risks related to privacy, impersonation, and misuse of synthetic 3D assets. All models used in this work are trained on publicly available datasets and follow the usage policies of those datasets.
To promote transparency and responsible research, we will release the implementation details necessary to reproduce our results. We encourage researchers and practitioners to use \ourmethod responsibly and to consider the broader societal implications when deploying image-to-3D synthesis .

\minisection{Reproducibility Statement.}
To ensure reproducibility, we will release the source code and training scripts required to reproduce the experimental results reported in this paper after the peer review process. All experiments are conducted using publicly available datasets, and detailed descriptions of the model architecture, training configuration, and evaluation procedures are provided in the main paper and appendices.

\minisection{Inference efficiency}
Table~\ref{tab:inference_time} compares the average inference
time per sample. Our method is the fastest among the evaluated
methods, requiring 21.96 seconds per sample, a 13.5\% reduction
compared to the fastest baseline, SADis.

\begin{table}[h]
\centering
\caption{\textbf{Inference time comparison.}
Average runtime in seconds per sample. Lower is better.}
\vspace{3mm}
\label{tab:inference_time}
\setlength{\tabcolsep}{6pt}
\begin{tabular}{lcccc}
\toprule
\textbf{Method} & MorphAny3D & SADis & StyleSculptor & \textbf{Ours} \\
\midrule
Time (s) $\downarrow$ & 64.70 & 25.39 & 39.76 & \textbf{21.96} \\
\bottomrule
\end{tabular}
\end{table}

\clearpage

\section{VLM-based Evaluation Protocol}
\label{appendix:vlm_eval}
\vspace{-2mm}

\minisection{Overview.}
To quantitatively evaluate the quality of generated images, we employ the InternVL2.5-8B~\cite{chen2024internvl25} model as an automatic evaluator. The vision-language model (VLM) performs binary judgments on two complementary aspects:

\begin{itemize}
\item \textbf{Style Consistency}: structural similarity between images while being invariant to color differences.
\item \textbf{Color Fidelity}: whether the generated image faithfully preserves the reference color distribution without introducing extraneous colors.
\end{itemize}

This enables semantic-level evaluation without requiring human annotation.

\minisection{Input Construction.}
Given a generated image $I_{\text{gen}}$, we construct paired inputs with different references:
\begin{itemize}
    \item $I_{\text{txu}}$: texture reference image
    \item $I_{\text{col}}$: color reference image
\end{itemize}

Each image is converted into a sequence of visual tokens (patches) using a vision encoder. Let:
\begin{equation}
    V_{\text{gen}}, \; V_{\text{txu}}, \; V_{\text{col}}
\end{equation}

denote the corresponding token representations.
We concatenate tokens as:
\begin{equation}
    V = [V_{\text{ref}}, V_{\text{gen}}]
\end{equation}
and provide the number of patches per image to preserve image boundaries:
\begin{equation}
    \texttt{num\_patches\_list} = [|V_{\text{ref}}|, |V_{\text{gen}}|]
\end{equation}

\minisection{Style Consistency Evaluation.}
To assess whether the generated image preserves structural/style characteristics, we prompt the VLM with:
\begin{quote}
\small
\texttt{Image-1: <Texture image> \\
Image-2: <Generation image> \\
Ignoring color differences, are the structures or styles of these two images similar? Answer yes or no.}
\end{quote}
The input pair is $(I_{\text{txu}}, I_{\text{gen}})$, and the model outputs:
\begin{equation}
    y_{\text{style}} \in \{\text{yes}, \text{no}\}
\end{equation}
We define the style consistency score over $N$ samples as:
\begin{equation}
    S_{\text{style}} = \frac{1}{N} \sum_{i=1}^{N} \mathbb{1}\left(y_{\text{style}}^{(i)} = \text{yes}\right)
\end{equation}

\minisection{Color Novelty Evaluation.}
To evaluate whether new colors are introduced, we use the prompt:
\begin{quote}
\small
\texttt{Image-1: <Color image> \\
Image-2: <Generation image> \\
If the second image shows a color that is not present in the first image, answer yes; otherwise, answer no.}
\end{quote}

The input pair is $(I_{\text{col}}, I_{\text{gen}})$, and the output is:
\begin{equation}
    y_{\text{color}} \in \{\text{yes}, \text{no}\}
\end{equation}

We compute the color novelty score as:
\begin{equation}
S_{\text{color}} = \frac{1}{N} \sum_{i=1}^{N} \mathbb{1}\left(y_{\text{color}}^{(i)} = \text{no}\right)
\end{equation}

\section{Hyperparameter visualization}
\label{appendix:lamda_vis}
\vspace{-2mm}

\begin{figure}[t]
    \centering
    \includegraphics[width=0.999\textwidth]{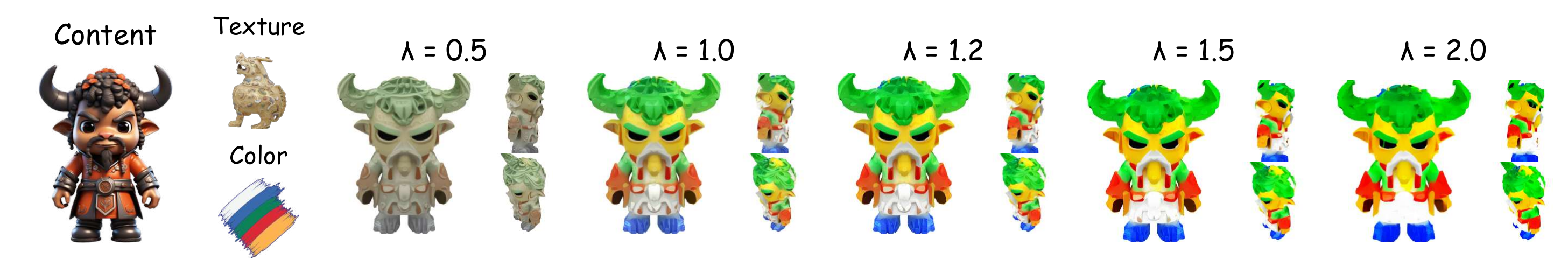}
    \vspace{-5mm}
    \caption{Ablation on $\lambda$.}
    \vspace{-3mm}
    \label{fig:ablation_lamda_vis}
\end{figure}

We further analyze the influence of the hyperparameter $\lambda$ in Equ.~\ref{eq:sa_fusion} on disentangled generation quality. Specifically, $\lambda$ controls the contribution of the color branch during feature fusion. We evaluate five settings, $\lambda \in {0.5, 1.0, 1.2, 1.5, 2.0}$, and visualize the corresponding results in Fig.~\ref{fig:ablation_lamda_vis}.
As shown in the figure, smaller values of $\lambda$ (e.g., $0.5$) lead to insufficient color guidance, causing the generated 3D assets to inherit excessive appearance information from the texture branch. In contrast, excessively large values of $\lambda$ (e.g., $1.5$ or $2.0$) overemphasize color features, which weakens texture fidelity and introduces structural inconsistencies.
We observe that $\lambda = 1.2$ achieves the best balance between texture preservation and color controllability. Under this setting, the generated results faithfully maintain the structural patterns from the texture reference while accurately matching the target color distribution, leading to the most visually coherent disentangled generation results.

\section{Ablation on k}
\begin{figure}[t]
    \centering
    \includegraphics[width=0.999\textwidth]{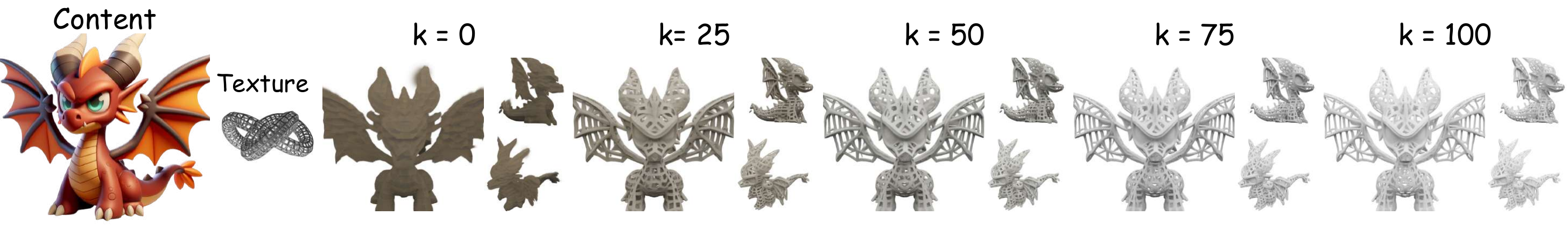}
    \vspace{-5mm}
    \caption{Ablation on $k$.}
    \vspace{-5mm}
    \label{fig:ablation_k_vis}
\end{figure}

\vspace{-2mm}
We conduct an ablation study by varying $k \in \{0, 25, 50, 75, 100\}$, where only the top-$k$ channels are retained and the features of the remaining channels are set to zero.
Fig.~\ref{fig:ablation_k_vis} shows that even when $k=25$, the generated 3D assets already exhibit clearly recognizable texture patterns, indicating that a relatively small subset of channels is sufficient to preserve most of the texture information, which further demonstrates the overcompleteness of the TRELLIS feature space.

\label{appendix:k_vis}

\section{Additional Visualization Results}
\label{appendix:supp_compares_vis}
\vspace{-2mm}
\begin{figure}[t]
    \centering
    \includegraphics[width=0.95\textwidth]{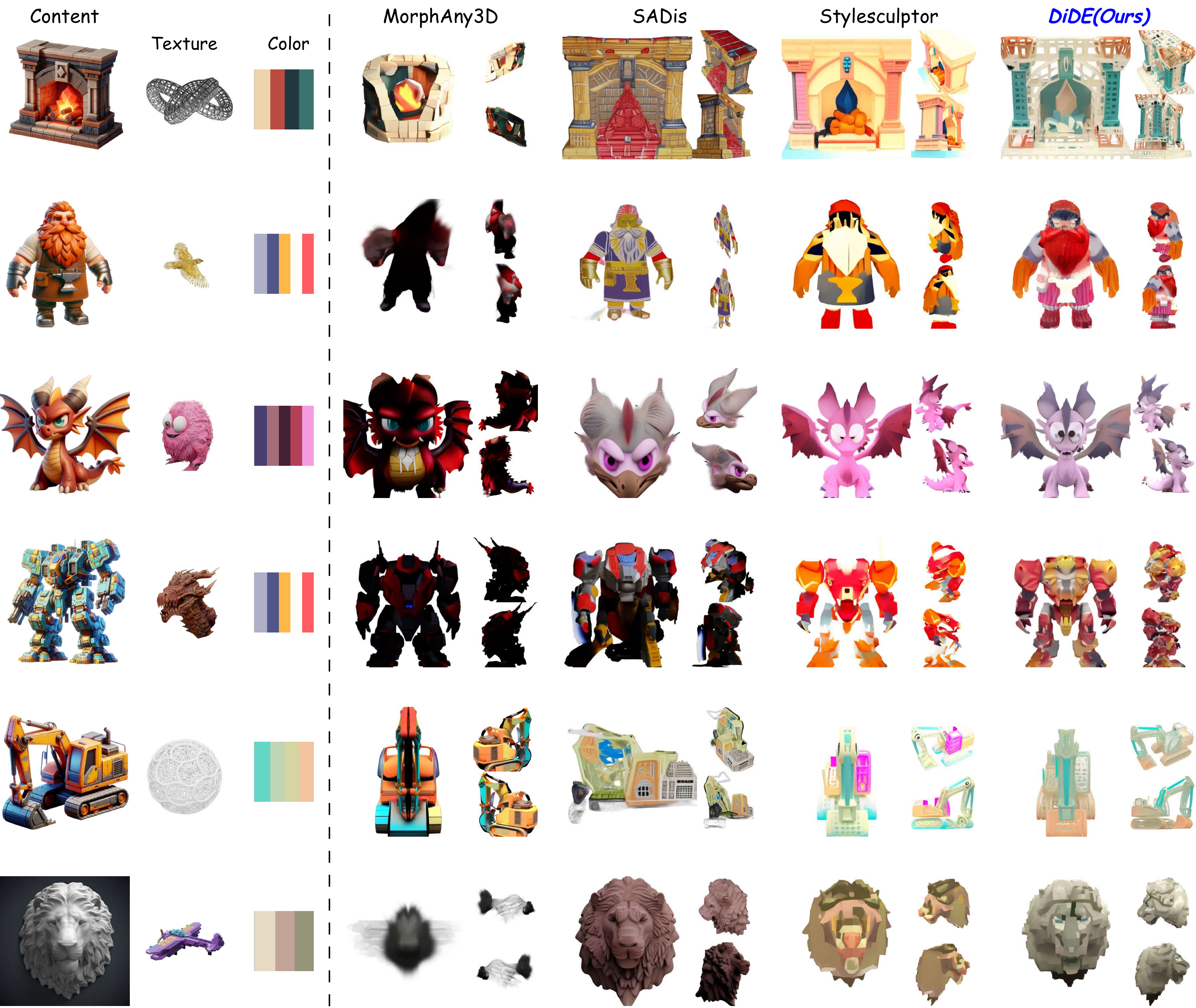}
    \caption{Additional qualitative comparisons}
    \label{fig:supp_compares}
\end{figure}

we provide additional qualitative results to further demonstrate the effectiveness and robustness of our method across diverse texture and color references. As shown in Fig.~\ref{fig:supp_compares}, our method consistently achieves effective disentanglement between texture structure and color attributes under various challenging scenarios.
Specifically, the generated 3D assets faithfully preserve the structural patterns and geometric details from the texture reference while accurately matching the target color distribution from the color reference image. Moreover, our method maintains strong content consistency without introducing noticeable distortions, structural artifacts, or undesired color leakage.
These additional visualizations further validate the generalization capability and stability of \ourmethod for controllable disentangled 3D generation.

\section{PCA Reconstruction Analysis}
\label{appendix:pca}
\vspace{-2mm}
\begin{figure}[t]
\centering
\includegraphics[width=0.7\textwidth]{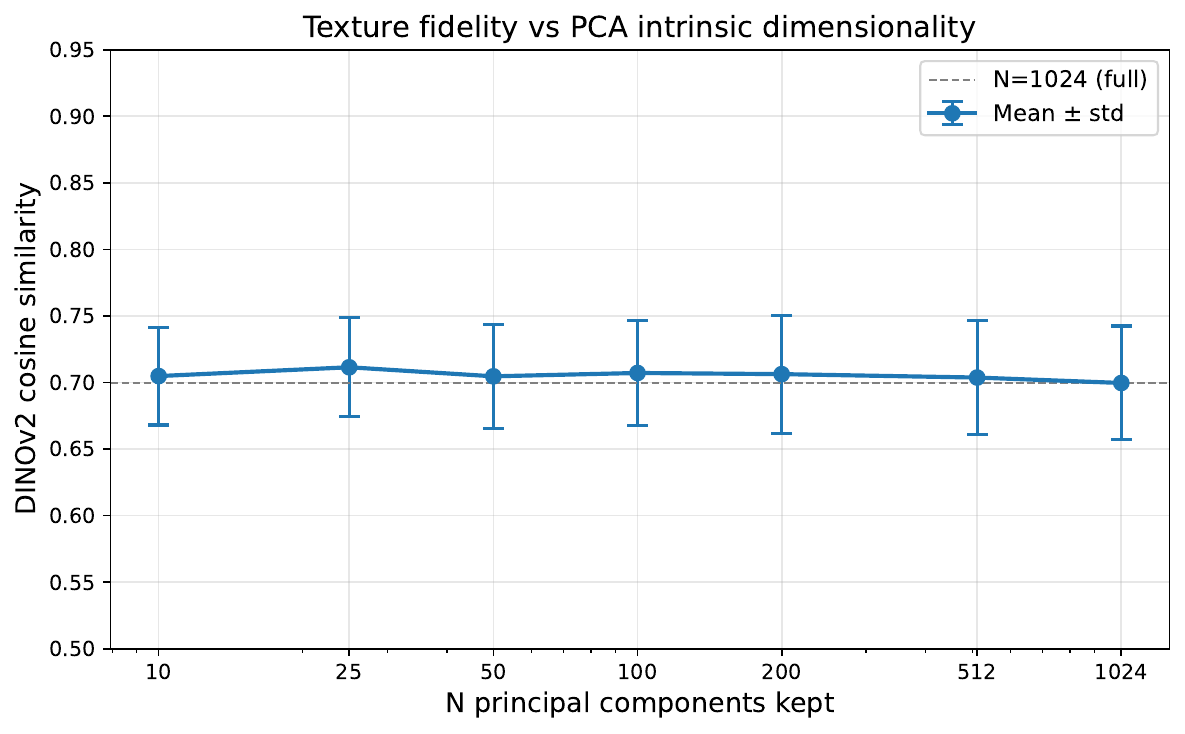}
\caption{DINOv2 cosine similarity between rendered outputs and texture references as a function of the number of PCA components retained from TRELLIS  texture-significant features. Similarity stabilizes after as few as 10 components, confirming the low intrinsic dimensionality of texture within the feature space.}
\vspace{-5mm}
\label{fig:pca_analysis}
\end{figure}
To further validate the overcompleteness claim of Sec.~\ref{sec:overcompleteness}, we perform a PCA analysis on the style-significant channels of the TRELLIS Stage II features. We project the style branch features onto an increasing number of principal components and measure DINOv2 cosine similarity between the rendered output and the texture reference.
As shown in Fig.~\ref{fig:pca_analysis}, similarity stabilizes after retaining as few as 10 principal components, despite the full channel dimensionality being much larger. This confirms that texture occupies an extremely low-dimensional subspace, leaving the majority of the feature space free to be independently repurposed for color encoding — directly motivating our channel partition design.

\clearpage

\end{document}